\documentclass[sigconf]{acmart}
\usepackage{float}
\usepackage{graphicx}
\usepackage{algorithm}
\usepackage[noend]{algpseudocode}
\usepackage{enumitem} % For customizing list environments
\usepackage{booktabs}
\usepackage{array}
\usepackage{multirow}
\usepackage{makecell}
\usepackage{bbding}
\usepackage{amsmath}

\usepackage{listings}
\usepackage{xcolor}
\usepackage{tcolorbox}
\tcbuselibrary{listings}% 使用 listings 库

\AtBeginDocument{%
  }

\setcopyright{acmlicensed}
\copyrightyear{2018}
\acmYear{2018}
\acmDOI{XXXXXXX.XXXXXXX}
\acmConference[Conference acronym 'XX]{Make sure to enter the correct
  conference title from your rights confirmation email}{June 03--05,
  2018}{Woodstock, NY}
\acmISBN{978-1-4503-XXXX-X/2018/06}

\begin{document}
\settopmatter{printacmref=false}
\renewcommand\footnotetextcopyrightpermission[1]{}

\title{PTCL: Pseudo-Label Temporal Curriculum Learning for Label-Limited Dynamic Graph}

\newcommand{\ourmethod}[1]{\texttt{PTCL}}
\author{Shengtao Zhang}
\authornote{Both authors contributed equally to this research.}
\email{zhangst@stu.xjtu.edu.cn}
\affiliation{%
  \institution{Xi’an Jiaotong University}
  \city{Xi'an}
  \state{Shaanxi}
  \country{China}
}

\author{Haokai Zhang}
\authornotemark[1]
\email{zhanghaokai@stu.xjtu.edu.cn}
\affiliation{%
  \institution{Xi’an Jiaotong University}
  \city{Xi'an}
  \state{Shaanxi}
  \country{China}
}

\author{Shiqi Lou}
\email{shiqilou01@gmail.com}
\authornote{Both authors contributed equally to this research.}
\affiliation{%
  \institution{Xi’an Jiaotong University}
  \city{Xi'an}
  \state{Shaanxi}
  \country{China}
}

\author{Zicheng Wang}
\authornotemark[2]
\email{cicadidae@stu.xjtu.edu.cn}
\affiliation{%
  \institution{Xi’an Jiaotong University}
  \city{Xi'an}
  \state{Shaanxi}
  \country{China}
}

\author{Zinan Zeng}
\email{2194214554@stu.xjtu.edu.cn}
\affiliation{%
  \institution{Xi’an Jiaotong University}
  \city{Xi'an}
  \state{Shaanxi}
  \country{China}
}

\author{Yilin Wang}
\email{13148035071xjtu@stu.xjtu.edu.cn}
\affiliation{%
  \institution{Xi’an Jiaotong University}
  \city{Xi'an}
  \state{Shaanxi}
  \country{China}
}

\author{Minnan Luo}
\email{minnluo@xjtu.edu.cn}
\authornote{Corresponding author: Minnan Luo, School of Computer Science and Technology,
Xi’an Jiaotong University, Xi’an 710049, China.}
\affiliation{%
  \institution{Xi'an Jiaotong University}
  \city{Xi'an}
  \state{Shaanxi}
  \country{China}
}

\renewcommand{\shortauthors}{Trovato et al.}

\begin{abstract}
Dynamic node classification is critical for modeling evolving systems like financial transactions and academic collaborations.
In such systems, dynamically capturing node information changes is critical for dynamic node classification, which usually requires all labels at every timestamp. However, it is difficult to collect all dynamic labels in real-world scenarios due to high annotation costs and label uncertainty (e.g., ambiguous or delayed labels in fraud detection). In contrast, final timestamp labels are easier to obtain as they rely on complete temporal patterns and are usually maintained as a unique label for each user in many open platforms, without tracking the history data. To bridge this gap, we propose \ourmethod{} (\textbf{P}seudo-label \textbf{T}emporal \textbf{C}urriculum \textbf{L}earning), a pioneering method addressing label-limited dynamic node classification where only final labels are available. \ourmethod{} introduces: (1) a temporal decoupling architecture separating the backbone (learning time-aware representations) and decoder (strictly aligned with final labels), which generate pseudo-labels, and (2) a Temporal Curriculum Learning strategy that prioritizes pseudo-labels closer to the final timestamp by assigning them higher weights using an exponentially decaying function. We contribute a new academic dataset (CoOAG), capturing long-range research interest in dynamic graph. Experiments across real-world scenarios demonstrate \ourmethod{}’s consistent superiority over other methods adapted to this task. Beyond methodology, we propose a unified framework FLiD (\textbf{F}ramework for \textbf{L}abel-L\textbf{i}mited \textbf{D}ynamic Node Classification), consisting of a complete preparation workflow, training pipeline, and evaluation standards, and supporting various models and datasets. The code can be found at \url{https://github.com/3205914485/FLiD}.
\end{abstract}






\maketitle{}

\section{Introduction}

Graph-structured data has become increasingly prevalent in various domains, ranging from social networks \cite{ying2018graph, newman2002random, feng2022twibot} and biological systems \cite{zitnik2018modeling, li2022graph} to financial transactions \cite{huang2022dgraph, li2023research} and academic collaborations \cite{hu2021ogb, zhou2022tgl}. Graphs provide a natural way to represent relationships and interactions between entities, making them a powerful tool for modeling complex systems \cite{bronstein2017geometric}. In particular, node classification, a fundamental task in graph analysis, aims to assign labels to nodes based on their features and the structure of the graph \cite{xiao2022graph, bhagat2011node, rong2019dropedge}. This task has been extensively studied in static graphs, where the graph structure and node features remain unchanged over time.

Nowadays, many real-world graphs are inherently dynamic, with nodes and edges evolving over time \cite{rossi2020temporal, kumar2019predicting, xu2020inductive}. For instance, in academic collaboration networks, authors' research interests may shift as they publish new papers, leading to changes in their labels (e.g., from "computer vision" to "natural language processing") \cite{jia2017quantifying}. Similarly, in financial transaction networks, users' behavior may change over time, potentially transitioning from "normal" to "fraudulent" activities \cite{huang2022dgraph}. These scenarios highlight the importance of dynamic label node classification, where the goal is to classify nodes in a dynamic graph whose labels may change over time.
\begin{figure}[t]
    \centering
    \includegraphics[width=1.0\linewidth]{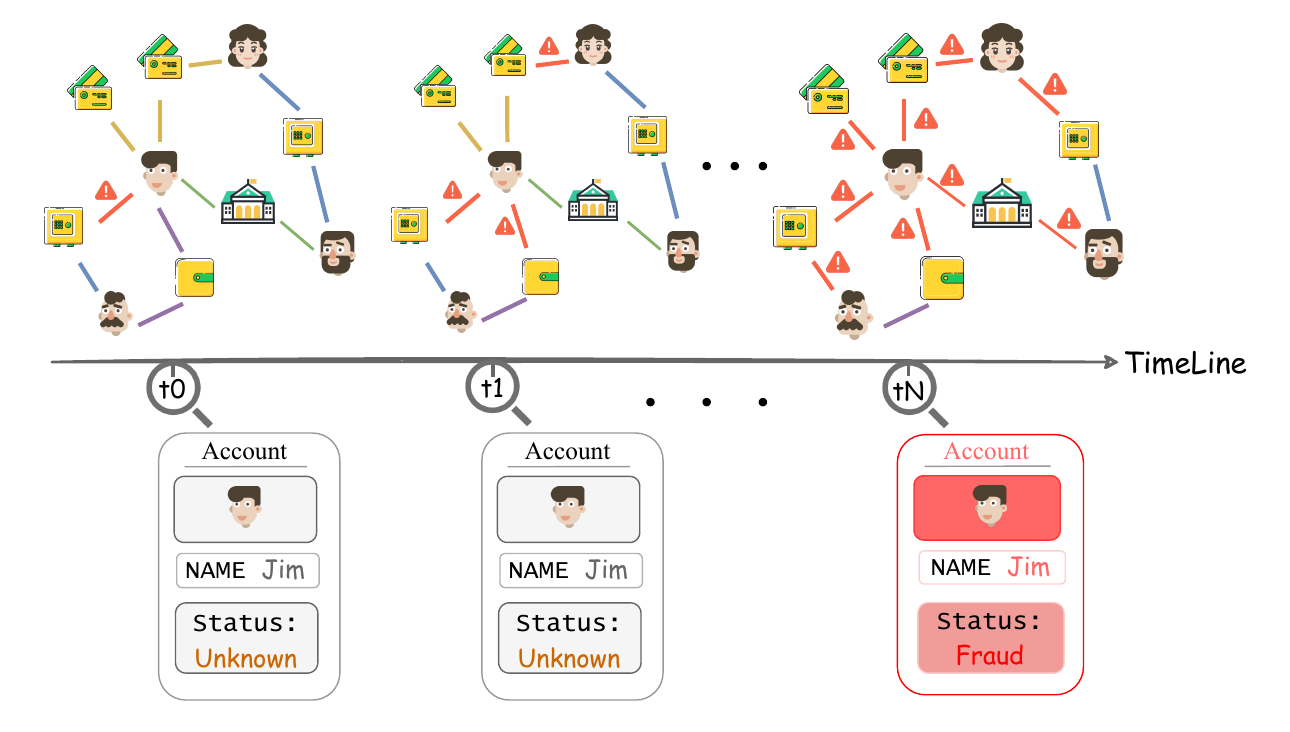}
    \caption{A present of a financial system. The graph represents a dynamic financial system where nodes represent entities such as users, payment cards, and financial institutions, while edges represent transactional relationships. Over time, user behavior is tracked through a sequence of transactions, and some users' labels(account status) may eventually be identified as fraudulent.}
    \label{fig:intro}
    \vspace{-12pt}
\end{figure}

 Ideally, we can get an experienced classifier through training with a highly dynamically changing trajectory. However, the challenge in dynamic graph node classification arises from the difficulty of collecting dynamic node labels due to high annotation costs (e.g., requiring continuous monitoring and manual effort) and label uncertainty (e.g., ambiguous or delayed labels in fraud detection). Although some dynamic datasets exist, they often provide weak dynamic labels that rarely change \cite{kumar2019predicting}, making it difficult to capture nodes' dynamic evolution effectively. While collecting labels at every timestamp is challenging, obtaining a single label at the final timestamp is relatively straightforward. As shown in Figure \ref{fig:intro}, in a financial transaction network, it may be easier to determine whether a user is fraudulent at the end of a given period rather than monitoring their behavior at each time step \cite{huang2022dgraph}. And many open platforms, such as OAG \cite{sinhaOverviewMicrosoftAcademic2015,zhangOAGLinkingLargescale2019,zhangOAGLinkingEntities2023,tangArnetMinerExtractionMining2008}, maintain a single final label for each user, like assigning each author a fixed research interest without tracking changes over time.

This observation aligns with the core challenge of our task, \textbf{label-limited dynamic node classification}. The objective is to classify nodes in dynamic graphs with limited label information, particularly focusing on the final timestamp labels, as these labels inherently capture the comprehensive status of nodes over time. To effectively leverage the dynamic information from previous timestamps, which lack labels, it is crucial to develop a robust model to represent and utilize these historical data.

One naive solution is to use the final timestamp labels as a substitute for the node's label at previous timestamps. Specifically, we can replicate the final labels across all previous timestamps, treating the node's label as static up to the final timestamp. However, this approach fails to capture the true temporal evolution of node labels, causing significant approximation errors.

The challenge associated with dynamic label collection naturally lead us to consider \textbf{semi-supervised learning (SSL)} \cite{yang2022survey,zhou2020time,zhu2005semi,learning2006semi,chapelle2006introduction,van2020survey} as a potential solution. A common SSL approach, \textbf{pseudo-labeling} \cite{lee2013pseudo,kage2024review}, can directly generate temporary labels for unlabeled data based on model predictions, and has been widely used in graph learning \cite{Li2022InformativePF, Li2018NaiveSD, Li2018DeeperII, Sun2019MultiStageSL}. In the context of our task, pseudo-labeling can be employed to generate labels of earlier timestamps, enabling the model to capture the underlying structure and dynamics of the graph while iteratively refining its predictions.

Building on the foundation of SSL and pseudo-labeling, we propose a novel method \textbf{\ourmethod{}} (\textbf{P}seudo-label \textbf{T}emporal \textbf{C}urriculum \textbf{L}earning). Our model includes a backbone and a decoder. The backbone is designed to model the dynamic graph data, while the decoder predicts node labels from backbone embeddings \cite{yu2023towards, rossi2020temporal}. To align the decoder with final timestamp labels while leveraging earlier timestamps for capturing node label evolution, we draw inspiration from prior works on the variational EM (Expectation Maximization) framework \cite{qu2024towards,Qu2019GMNNGM,Zhao2022LearningOL,Neal1998AVO,Dempster1977MaximumLF}, which separates the optimization of model components to iteratively refine predictions. Specifically, we train the decoder exclusively on the final timestamp labels, to ensure the consistency with the only available labels. Once the decoder is trained, it generates pseudo-labels for all timestamps, which serve as approximations of the true labels at earlier time steps. These pseudo-labels, along with the final timestamp labels, are then used to train the backbone via a weighted loss function, allowing the backbone to learn embeddings that capture the temporal evolution of node labels.

To address the potential issue of low-quality pseudo-labels during the initial stages of EM iteration, inspired by \textbf{Curriculum Learning} \cite{Bengio2009CurriculumL, Wang2021ASO, Soviany2021CurriculumLA}, we design a Temporal Curriculum Learning method to compute temporal weights for pseudo-labels. Based on the easy-to-hard rule, we assign higher weights to pseudo-labels near the final timestamps in early training, as they are easier to predict. As training progresses, we gradually shift focus to earlier timestamps, guiding the model toward more challenging examples.

Extensive experiments show that \ourmethod{} consistently outperforms other methods adapted to this task across multiple datasets, validating its effectiveness in capturing the temporal evolution of nodes. Additionally, we conduct a series of additional studies to verify the contribution of each design of \ourmethod{}.

To sum up, our contributions are as follows:
\begin{itemize}[left=0pt..1em, topsep=0pt, itemsep=-1pt, parsep=0pt]
    \item \textbf{Pioneering Study}: To the best of our knowledge, this is a pioneering study to systematically investigate the problem of label-limited dynamic node classification. We formalize the task, identify its unique challenges, and propose a comprehensive method to address them.

    \item \textbf{Novel Method}: We introduce a new method \ourmethod{}, which captures the dynamic nature of nodes with limited labels, advancing dynamic graph study by constructing highly varying history information. Various experiments demonstrate the effectiveness of \ourmethod{} and the necessity of each design.
   
    \item \textbf{New Dataset}: We contribute a new dataset, CoOAG,  which is derived from academic collaboration networks and designed for dynamic graph learning. It captures the dynamic nature of research interests, providing a rich testbed for evaluating \ourmethod{}.

    \item \textbf{Unified Framework}: We propose a unified framework, FLiD (\textbf{F}ramework for \textbf{L}abel-L\textbf{i}mited \textbf{D}ynamic Node Classification), for our task, which includes a complete preparation workflow, a training pipeline, and evaluation protocols. Our framework supports various backbones and datasets, offering a flexible and extensible solution.

\end{itemize}

\section{Related Work}
\label{related_work}
\subsection{Dynamic Node Classification}

Dynamic graph learning is typically divided into discrete-time and continuous-time approaches. Discrete-time methods segment dynamic graphs into snapshot sequences \cite{fan2021gcn,sankar2020dysat}, whereas continuous-time methods capture fine-grained temporal interactions for better real-world alignment \cite{rossi2020temporal,Wang2021TCLTD}. While link prediction dominates existing research in downstream tasks \cite{qin2023temporal,yu2023towards}, dynamic node classification remains underexplored despite its practical significance.

Existing works have explored various approaches to dynamic node classification: JODIE \cite{kumar2019predicting} predicts user state changes in user-item graphs, DynPPE \cite{guo2021subset} classifies nodes in large-scale dynamic graphs using a discrete-time method, and OTGNet \cite{feng2023towards} handles dynamically expanding class sets. However, these methods overlook the underlying node dynamics and assume full dynamic labels, which rarely hold in real-world scenarios. We address the critical yet understudied problem of label-limited dynamic node classification, where only final timestamp annotations are available. To capture node dynamics effectively, we propose \ourmethod{}, a continuous-time method leveraging pseudo-labels with Curriculum Learning.

\subsection{Pseudo-Labeling}
\label{Related Work Pseudo-Labeling}
Pseudo-labeling \cite{lee2013pseudo} has emerged as a prominent and widely utilized technique in semi-supervised learning, primarily aimed at predicting labels for unlabeled data. This approach functions as a form of entropy minimization, which encourages the reduction of data embedding density near decision boundaries, thereby facilitating the formation of more robust boundaries in low-density regions of the embedding space \cite{Pei2024MemoryDA,Chapelle2005SemiSupervisedCB,Grandvalet2004SemisupervisedLB}. Its has been applied across various domains \cite{kage2024review}, including computer vision \cite{lee2013pseudo, xu2021end}, graph learning \cite{Li2022InformativePF, Li2018NaiveSD, Li2018DeeperII, Sun2019MultiStageSL}, knowledge distillation \cite{Hinton2015DistillingTK}, and adversarial training \cite{Miyato2017VirtualAT,Xie2019UnsupervisedDA}.

However, the efficiency of pseudo-labeling highly relies on the accuracy of the pseudo-labels. Incorrect pseudo-labels can severely impair model performance by introducing noise and misleading patterns into the learning process \cite{Pei2024MemoryDA}. A natural strategy to address this issue is Curriculum Learning \cite{Bengio2009CurriculumL}, which involves initially training the model on "easier" examples of a class or concept, followed by progressively more challenging ones. This ensures that the model first assigns correct pseudo-labels to the easier examples from the unlabeled set, thereby reliably expanding the labeled dataset. Building on this foundation, numerous studies have proposed various metrics to identify easier examples and more reliable pseudo-labels \cite{kage2024review}. For instance, \citet{CascanteBonilla2020CurriculumLR}, \citet{He2022HowDP}, and \citet{Sun2019MultiStageSL} employ a \textit{confidence score threshold}, while \citet{Pei2024MemoryDA} and \citet{Song2019SELFIERU} utilize the \textit{entropy of softmax trajectory} over previous epochs. 
% Additionally, \citet{Zhang2021FlexMatchBS} enhances the confidence score threshold by introducing \textit{dynamically-selected thresholds per-class}, which are adjusted based on the perceived difficulty of learning each class's label to avoid confirmation bias \cite{Arazo2019PseudoLabelingAC}.

In the context of our research, existing methods overlook the temporal dynamics inherent in the evolving nature of graphs. In contrast, our proposed Temporal Curriculum Learning explicitly integrates temporal considerations, allowing the model to adapt more effectively to shifts in data distribution and decision boundaries as learning progresses.

\subsection{Variational EM Framework}
The variational EM framework \cite{Dempster1977MaximumLF,Neal1998AVO} is a widely used framework for parameter estimation in probabilistic models with latent variables. In the classical EM algorithm, the goal is to maximize the likelihood of observed data by iteratively refining model parameters through alternating E-steps (expectation computation) and M-steps (parameter maximization). GMNN \cite{Qu2019GMNNGM} applies EM for semi-supervised static graphs and GLEM \cite{Zhao2022LearningOL} combines GNNs with language models. Our contribution lies not in framework innovation but in adapting this established paradigm to address label-limited dynamic node classification.

% \subsection{Variational EM Framework}
% The variational EM framework \cite{Dempster1977MaximumLF,Neal1998AVO} is a widely used framework for parameter estimation in probabilistic models with latent variables. In the classical EM algorithm, the goal is to maximize the likelihood of observed data by iteratively refining model parameters. The algorithm alternates between two steps: the E-step, which computes the expected value of the complete-data log-likelihood given the observed data and current parameter estimates, and the M-step, which updates the model parameters to maximize this expected log-likelihood. This process implicitly handles hidden variables by marginalizing their possible values during the E-step.
% GMNN \cite{Qu2019GMNNGM} employs two GNNs to model label dependencies and approximate posteriors via variational EM for semi-supervised node classification. GLEM \cite{Zhao2022LearningOL} integrates graph neural networks (GNNs) and language models (LMs) under a variational EM framework for text-attributed graphs. 

\section{Problem Formulation}
A \textbf{dynamic graph with dynamic labels} can be mathematically represented as a sequence of chronologically ordered events:
\(\mathcal{G} = \{x(t_i)\} = \{(u_i, v_i, t_i)\}\), 
where $0 \leq t_1 \leq t_2 \leq \cdots$.   
Each event $x(t_i)$ describes an interaction between a source node $u_i \in \mathcal{V}$ and a destination node $v_i \in \mathcal{V}$ at time $t_i$. And $y_{u_i}^{t_i}, y_{v_i}^{t_i} \in \mathcal{Y}$ are their respective labels at \(t_i\). 
$\mathcal{V}$ denotes the set of all nodes, and $\mathcal{Y}$ is the class set of all nodes.
For each node $u \in \mathcal{V}$, \(\mathcal{T}_u = \{ t_i \mid u = u_i \text{ or } u = v_i \text{ in } x(t_i) \in \mathcal{G} \}\) is the set of all timestamps at which $u$ participates in any event in $\mathcal{G}$. 
The last occurrence time \(T_u = \max \mathcal{T}_u\) is the most recent timestamp in $\mathcal{T}_u$. 
We further define $\mathcal{Y}_F = \{y_u^{T_u} \mid u \in \mathcal{V}\}$, the set of ground-truth labels at the final timestamps, and $\mathcal{Y}_E = \{y_u^t \mid u \in \mathcal{V}, t \in \mathcal{T}_u \setminus \{T_u\}\}$, the set of labels at non-last timestamps of every node. In most cases, $|\mathcal{Y}_E| \gg |\mathcal{Y}_F|$.
In our research scenario, $\mathcal{Y}_F$ are known, whereas $\mathcal{Y}_E$ are considered unknown due to data collection constraints.
Following the training-evaluation paradigm, we determine a boundary time $T_B$ to separate the training and evaluation datasets. The final timestamp label set $\mathcal{Y}_F$ is then divided into two subsets: \(\mathcal{Y}_{F,B} = \{y_u^{T_u} \mid u \in \mathcal{V}, T_u \leq T_B\}\) consisting of labels for nodes whose final timestamps are before $T_B$, and \(\mathcal{Y}_{F,A} = \{y_u^{T_u} \mid u \in \mathcal{V}, T_u > T_B\}\) containing labels for nodes whose final timestamps is after $T_B$. Similarly, $\mathcal{Y}_E$ is partitioned into \(\mathcal{Y}_{E,B} = \{y_u^t \mid u \in \mathcal{V}, t \in \mathcal{T}_u \setminus \{T_u\}, t \leq T_B\}\) and \(\mathcal{Y}_{E,A} = \{y_u^t \mid u \in \mathcal{V}, t \in \mathcal{T}_u \setminus \{T_u\}, t > T_B\}\), representing labels whose corresponding timestamps are before and after $T_B$, respectively.

Specifically, when all timestamps $t_i$ are identical or omitted, the graph degenerates into a static graph \cite{kipf2016semi, holme2012temporal}, where each node $u \in \mathcal{V}$ is associated with a single label $y_u$. Then the problem degrades to a \textit{static node classification} task.
Alternatively, if labels are available for all nodes at all timestamps, i.e., $\mathcal{Y}_E$ is entirely known, the problem becomes a \textit{fully supervised dynamic node classification} task, which has been extensively studied in prior research \cite{xu2020inductive, rossi2020temporal, cong2023we, yu2023towards}.

The dynamic graph backbone generates node embeddings $\mathbf{h}_u^t$ for each node $u$ at each timestamp $t \in \mathcal{T}_u$. The backbone takes the node features as input  $\mathbf{n}_u \in \mathbb{R}^{d_N}$ and edge features $\mathbf{e}_{u,v}^t \in \mathbb{R}^{d_E}$.
% , which are collectively denoted as $\mathcal{A} = \{\mathbf{n}_u, \mathbf{e}_{u,v}^t \mid u, v \in \mathcal{V}, t \in \mathcal{T}_u\}$. Here, $\mathcal{A}$ represents the set of all node and edge features in the graph. 
If the graph is non-attributed, we assume $\mathbf{n}_u = \mathbf{0}$ and $\mathbf{e}_{u,v}^t = \mathbf{0}$ for all nodes and edges, respectively.

Given a dynamic graph $\mathcal{G}$ with dynamic labels and $|\mathcal{Y}_E| \gg |\mathcal{Y}_F|$, our task \textbf{label-limited dynamic node classification} aims to maximize $\log p(\mathcal{Y}_{F,B} | \mathcal{G})$. Specifically, our goal is to learn a model that can finally accurately predict $\mathcal{Y}_F$.

\begin{figure*}[t]
    \centering
    \includegraphics[width=\linewidth]{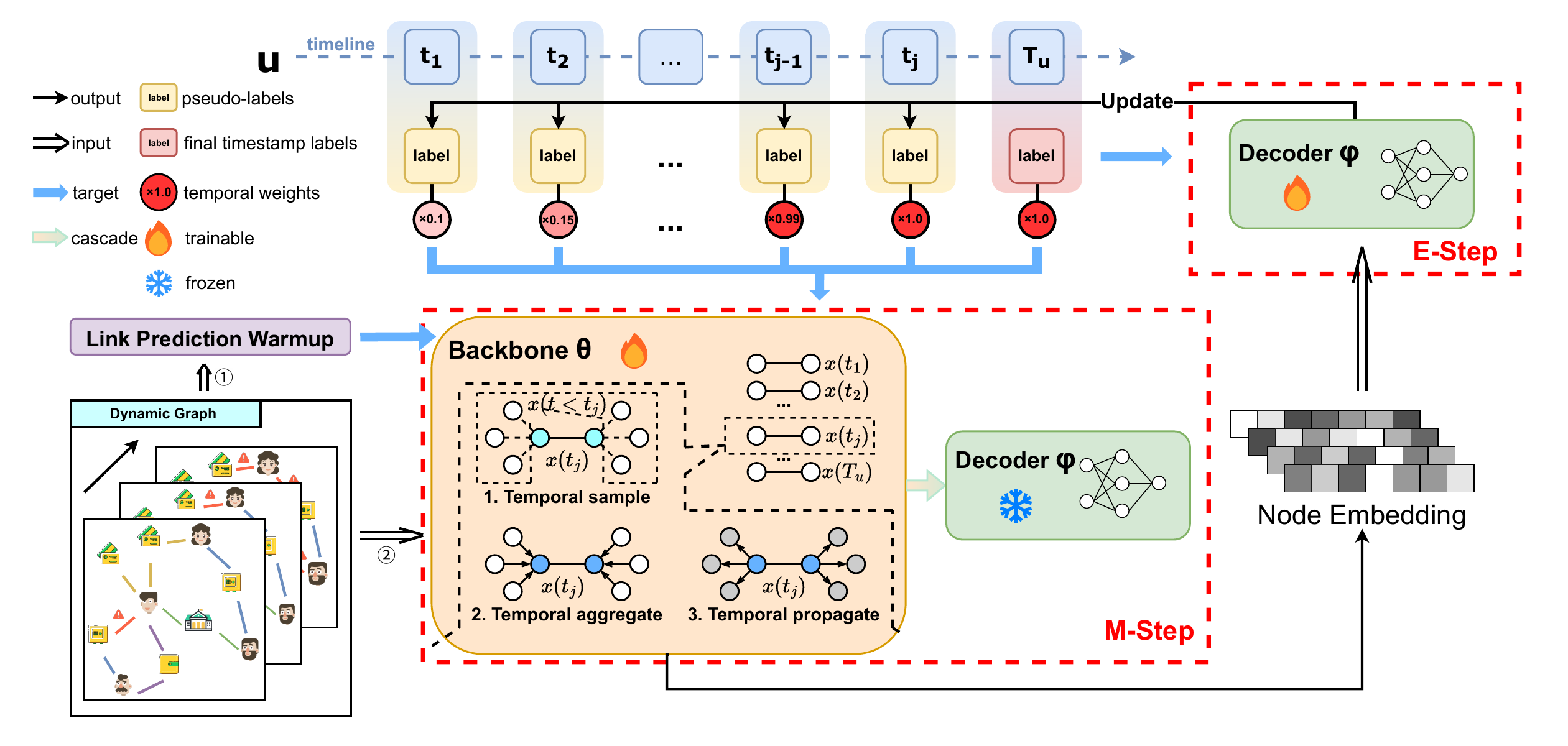}
    \caption{Overview of our proposed method. \ourmethod{} consists of a Variational EM process with a dynamic graph backbone and a decoder. During the warmup phase, the dynamic graph backbone is trained on a link prediction task, where the dynamic graph structure serves as the target. After warmup, in each M-step, the backbone receives final timestamp labels, pseudo-labels, and the dynamic graph structure as input, while the decoder is trained in the E-step to refine pseudo-labels. Additionally, the Temporal Curriculum Learning strategy prioritizes pseudo-labels based on their temporal proximity to the final timestamp labels, ensuring higher-quality training.}
    \label{overview}
\end{figure*}

\section{Methodology}
In this section, we present our proposed method for label-limited dynamic label learning. As shown in Figure \ref{overview}, our approach combines the variational EM framework with a novel Temporal Curriculum Learning strategy to effectively leverage both final timestamp labels and pseudo-labels in a dynamic graph setting. The EM framework alternates between two steps: the E-step, where the decoder is optimized to predict node labels using final timestamp labels, and the M-step, where the dynamic graph backbone is trained using both final timestamp labels and pseudo-labels generated by the decoder. To further enhance the quality of pseudo-labels, we introduce Temporal Curriculum Learning, which dynamically adjusts the weight of pseudo-labels based on their temporal distance to the final timestamp labels. This ensures that the backbone is trained with high-quality supervision signals.

\subsection{Variational EM Framework}
Following previous work \cite{Zhao2022LearningOL,Qu2019GMNNGM}, we adopt the variational EM framework \cite{Dempster1977MaximumLF,Neal1998AVO} to maximize $\log p(\mathcal{Y}_{F,B} | \mathcal{G})$.

\subsubsection{Evidence Lower Bound (ELBO)}
To handle the unknown labels $\mathcal{Y}_{E,B}$, instead of directly optimizing $\log p_{\theta}(\mathcal{Y}_{F,B} | \mathcal{G})$, we maximize the evidence lower bound (ELBO) of the log-likelihood function:
\begin{equation}
\begin{aligned}
    \log p(\mathcal{Y}_{F,B} | \mathcal{G}) \geq \mathbb{E}_{q_{\phi}(\mathcal{Y}_{E,B} | \mathcal{G})}[\log p_{\theta}(\mathcal{Y}_{F,B}, \mathcal{Y}_{E,B} | \mathcal{G}) \\ 
- \log q_{\phi}(\mathcal{Y}_{E,B} | \mathcal{G})],
\end{aligned}
\end{equation}
where $p_{\theta}(\mathcal{Y}_{F,B}, \mathcal{Y}_{E,B} | \mathcal{G})$ is the joint distribution of observed and unknown labels, modeled by the dynamic graph backbone with parameters $\theta$. $q_{\phi}(\mathcal{Y}_{E,B} | \mathcal{G})$ is the variational distribution that approximates the true posterior distribution $p_{\theta}(\mathcal{Y}_{E,B} | \mathcal{Y}_{F,B}, \mathcal{G})$, modeled by the decoder with parameters $\phi$.

To facilitate optimization, we follow mean field assumption  \cite{Getoor2001LearningPM}, which yields the following factorization:
\begin{equation}
q_{\phi}(\mathcal{Y}_{E,B} | \mathcal{G}) =  \prod_{u \in \mathcal{V}} \prod_{\substack{t \in \mathcal{T}_u \setminus \{T_u\} \\ t\leq T_B}} q_{\phi}(y_u^t | \mathbf{h}_u^t),
\end{equation}
% \vspace{-3pt}
where $q_{\phi}(y_u^t | \mathbf{h}_u^t)$ is the label distribution predicted by the decoder. It means the decoder maps the node embeddings $\mathbf{h}_u^t$ generated by the backbone to label predictions $\hat{y}_u^t$.

The ELBO can be optimized by alternating between the E-step and the M-step.

\subsubsection{Backbone Warmup}
Since the initial parameters of the EM algorithm are crucial for its performance \cite{dy2004feature, kwedlo2015new}, we first warm up the backbone by training it on a link prediction task,  which predicts the existence of edges between nodes based on their embeddings. Given a dynamic graph $\mathcal{G} = \{(u_i, v_i, t_i)\}$, the link prediction loss is defined as:
\begin{equation}
\begin{aligned}
    \mathcal{L}_{lp} = & -\sum_{(u_i, v_i, t_i) \in \mathcal{G}} \log \sigma(\text{MLP}(\mathbf{h}_{u_i}^{t_i} \parallel \mathbf{h}_{v_i}^{t_i})) \\
    & - \sum_{(u_j, v_j, t_j) \notin \mathcal{G}} \log \left(1 - \sigma(\text{MLP}(\mathbf{h}_{u_j}^{t_j} \parallel \mathbf{h}_{v_j}^{t_j}))\right).
\end{aligned}
\label{link prediction}
\end{equation}
where $\sigma(\cdot)$ is the sigmoid function, $\parallel$ means concatenation. The first term encourages the model to predict existing edges correctly, while the second term penalizes the model for predicting non-existent edges.

By minimizing $\mathcal{L}_{lp}$, the backbone learns to capture the structural and temporal dependencies in the graph, providing a strong initialization for the EM algorithm.

\subsubsection{E-step} \label{e-step}
In the E-step, we use the wake-sleep algorithm \cite{Hinton1995TheA}, following \citet{Zhao2022LearningOL}. We fix the dynamic graph backbone $\theta$ and optimize the decoder $\phi$ to minimize the KL divergence between the true posterior distribution $p_{\theta}(\mathcal{Y}_{E,B} | \mathcal{G}, \mathcal{Y}_{F,B})$ and the variational distribution $q_{\phi}(\mathcal{Y}_{E,B} | \mathcal{G})$. The objective function for the decoder is:
\begin{equation}
\begin{aligned}
\hat{\mathcal{O}_{\phi}} = \sum_{u \in \mathcal{V}} \sum_{\substack{t \in \mathcal{T}_u \setminus \{T_u\} \\ t\leq T_B}} \mathbb{E}_{p_{\theta}(y_u^t | \mathcal{G}, \mathcal{Y}_{F,B})}[\log q_{\phi}(y_u^t | \mathbf{h}_u^t)],
\end{aligned}
\end{equation}

Following \citet{Zhao2022LearningOL}, we use the pseudo-labels $\hat{\mathcal{Y}}_{E,B}$ generated by the decoder to approximate the distribution $p_{\theta}(y_u^t | \mathcal{G}, \mathcal{Y}_{F,B})$:
\begin{equation}
p_{\theta}(y_u^t | \mathcal{G}, \mathcal{Y}_{F,B}) \approx p_{\theta}(y_u^t | \mathcal{G}, \mathcal{Y}_{F,B}, \hat{\mathcal{Y}}_{E,B} \setminus \hat{\mathcal{Y}}_{E,B}^u),
\end{equation}
where \(\hat{\mathcal{Y}}_{E,B}^u = \{ \hat{y}_u^{t} \mid t \in \mathcal{T}_u \setminus \{T_u\}, t \leq T_B\}\) is the set of pseudo-labels of node $u$. Then the  objective function of the decoder changes to:
\begin{equation}
\label{e-objective1}
\begin{aligned}
\hat{\mathcal{O}}_{\phi} = \alpha & \sum_{u \in \mathcal{V}} \sum_{\substack{t \in \mathcal{T}_u \setminus \{T_u\} \\ t\leq T_B}} \mathbb{E}_{p_{\theta}(y_u^t | \mathcal{G}, \mathcal{Y}_{F,B})}[\log q_{\phi}(y_u^t | \mathbf{h}_u^t)]\\
+ & (1 - \alpha) \sum_{\substack{u \in \mathcal{V} \\ T_u \leq T_B}} \log q_{\phi}(y_u^{T_u} | \mathbf{h}_u^{T_u}),
\end{aligned}
\end{equation}
where $\alpha$ is a hyperparameter that balances the weight of pseudo-labels and final timestamp labels. 

But in practice, as shown in Section \ref{Necessity of Separate Optimization}, we find that setting $\alpha$ to 0 yields the best performance, which means we train the decoder only with final timestamp labels. Therefore, the final objective function for the decoder is: 
\begin{equation}
\mathcal{O}_{\phi} = \sum_{\substack{u \in \mathcal{V} \\ T_u \leq T_B}} \log q_{\phi}(y_u^{T_u} | \mathbf{h}_u^{T_u}),
\label{e-objective2}
\end{equation}

\subsubsection{M-step}
In the M-step, following the previous work \cite{Zhao2022LearningOL,Qu2019GMNNGM}, we aim to maximize the following pseudo-likelihood \cite{Besag1975StatisticalAO}. Specifically, we fix the decoder $\phi$ and optimize the dynamic graph backbone $\theta$ using both the final timestamp labels $\mathcal{Y}_{F,B}$ and the pseudo-labels $\hat{\mathcal{Y}}_{E,B}$ generated in the E-step. The objective is to maximize the pseudo-likelihood:
\begin{equation}
\begin{aligned}
\hat{\mathcal{O}}_{\theta} = \beta & \sum_{u \in \mathcal{V}} \sum_{\substack{t \in \mathcal{T}_u \setminus \{T_u\} \\ t \leq T_B}} \log p_{\theta}(y_u^t | \mathcal{G}, \mathcal{Y}_{F,B}, \hat{\mathcal{Y}}_{E,B} \setminus \hat{\mathcal{Y}}_{E,B}^u) \\
+ & (1 - \beta) \sum_{\substack{u \in \mathcal{V} \\ T_u \leq T_B}} \log p_{\theta}(y_u^{T_u} | \mathcal{G}, \mathcal{Y}_{F,B} \setminus \{y_u^{T_u}\}, \hat{\mathcal{Y}}_{E,B}),
\end{aligned}
\label{m-step objective}
\end{equation}
where $\beta$ is a hyperparameter that balances the weight of pseudo-labels and final timestamp labels. Note that the backbone can only generate embeddings, so we cascade the backbone and decoder, fixing the decoder $\phi$ to use final timestamp labels and pseudo-labels generated by the decoder to train the backbone.

\subsection{Temporal Curriculum Learning}
In the M-step, the backbone is trained using both final timestamp labels $\mathcal{Y}_{F,B}$ and pseudo-labels $\hat{\mathcal{Y}}_{E,B}$ generated by the decoder. However, the quality of pseudo-labels can vary significantly, as nodes' labels may change over time. To ensure that the backbone is trained with high-quality pseudo-labels, we propose a \textbf{Temporal Curriculum Learning} strategy, which dynamically adjusts the weight of pseudo-labels based on their temporal proximity to the final timestamp labels and $\tau$ (EM iteration counter).

\subsubsection{Motivation}
Temporal Curriculum Learning is motivated by the inherent continuity in node label evolution: a node's label $y_u^t$ at timestamp $t$ typically aligns more closely with its final label $y_u^{T_u}$ when $t$ nears $T_u$, while earlier timestamps exhibit greater divergence. This temporal proximity principle suggests that pseudo-labels near $T_u$ provide reliable supervision, whereas distant ones introduce higher noise. Our strategy thus prioritizes predictions near the final timestamp during the initial training iterations. As optimization progresses, the model gradually incorporates earlier pseudo-labels, leveraging their improving quality to capture comprehensive temporal evolution patterns.

\subsubsection{Pseudo-label Temporal Weighting}
To implement Temporal Curriculum Learning, we introduce a weighting mechanism for pseudo-labels based on their temporal order relative to the final timestamp $T_u$. Specifically, for each node $u \in \mathcal{V}$, timestamp $t \in \mathcal{T}_u$ in $\tau$-th iteration, we define a weight $w_u^{t,\tau}$ as follows:
\begin{align}
    & w_u^{t,\tau}=f_{\text{TW}}(t, u, \tau, T_u, \gamma) =
\begin{cases} 
1, & \text{if } d_u^t \leq \tau, \\
\exp\left(-\gamma  \cdot (d_u^t - \tau)\right), & \text{if } d_u^t > \tau,
\end{cases} \\
    & d_u^t = |\{ t' \in \mathcal{T}_u \mid t' > t \}|
\end{align}

where $d_u^t$ is the discrete temporal distance between of timestamp $t$ and $T_u$ in $\mathcal{T}_u$, representing its position in the sequence of timestamps for node $u$. For example, if $\mathcal{T}_u = \{t_1, t_2, t_3, T_u\}$ and $t_1 < t_2  < t_3 < T_u$, then $d_u^{t_1} = 3$\textbf{}, $d_u^{t_2} = 2$, $d_u^{t_3} = 1$ and $d_u^{T_u} = 0$. It serves as a threshold that determines whether a timestamp is considered "close" to the final timestamp $T_u$. $\gamma > 0$ is a hyperparameter that controls the rate of Temporal Curriculum Learning decay.

The weight $w_u^{t,\tau}$ dynamically adjusts the importance of pseudo-labels during training.  If $d_u^t \leq \tau$, the timestamp $t$ is considered close to the final timestamp $T_u$, and the pseudo-label is assigned a weight of 1, indicating high confidence in its quality. And if $d_u^t > \tau$, the timestamp $t$ is considered far from $T_u$, and the pseudo-label weight decays exponentially with the distance $\tau - d_u^t$, reducing its influence on the training process.

This mechanism ensures that the backbone is trained with higher-quality pseudo-labels during the early stages of training, while gradually reducing the impact of lower-quality pseudo-labels as training progresses. As $\tau$ increases over iterations, the model is able to learn from pseudo-labels at earlier timestamps, capturing the temporal evolution of node labels more effectively.

With the pseudo-label temporal weights $w_u^{t,\tau}$, the objective function for the M-step is modified as follows:
\begin{equation}
\begin{aligned}
\mathcal{O}_{\theta} = \beta & \sum_{u \in \mathcal{V}} \sum_{\substack{t \in \mathcal{T}_u \setminus \{T_u\} \\ t \leq T_B}} w_u^{t,\tau} \log p_{\theta}(y_u^t | \mathcal{G}, \mathcal{Y}_{F,B}, \hat{\mathcal{Y}}_{E,B} \setminus \hat{\mathcal{Y}}_{E,B}^u) \\
+ & (1 - \beta) \sum_{\substack{u \in \mathcal{V} \\ T_u \leq T_B}} \log p_{\theta}(y_u^{T_u} | \mathcal{G}, \mathcal{Y}_{F,B} \setminus \{y_u^{T_u}\}, \hat{\mathcal{Y}}_{E,B}).
\end{aligned}
\end{equation}
By incorporating temporal weight $w_u^{t,\tau}$, the backbone is trained to prioritize high-quality pseudo-labels.

\subsection{Learning and Optimization}
First, we use a link prediction task to warm up the backbone. Then we proceed with the variational EM algorithm, which alternates between the E-step and the M-step. Finally, we use the decoder to predict $\mathcal{Y}_{F,A}$. The complete procedure is summarized in Appendix. \ ref {algorithm}.

\subsection{FLiD: A novel framework}

We introduce \textbf{FLiD}, a novel code framework specifically designed for our task, where only final timestamp labels are available. Unlike existing dynamic graph frameworks such as DyGLib \cite{yu2023towards} and TGL \cite{Wang2021TCLTD}, FLiD provides comprehensive support through several key features. It includes new data preprocessing modules for handling raw datasets, and a flexible training pipeline adaptable to various methods, while also supporting different dynamic graph backbones and pseudo-label enhancement strategies. Additionally, FLiD incorporates specific evaluation protocols to ensure fair comparisons across methods. We use FLiD to implement \ourmethod{}, and all experiments in this study are conducted using FLiD, which is designed to be highly extensible, enabling seamless integration of future methods for this task.

\begin{table*}[htbp]
\centering
\caption{Performance comparison across datasets (Wikipedia, Reddit, Dsub, CoOAG). We run all experiments with five random seeds to ensure a consistent evaluation and report the average performance as well as the standard deviation in parentheses. Bold indicates the best performance, \underline{underline} the second best. Dsub and CoOAG datasets can not apply the DLS method due to a lack of dynamic labels. TGN runs out of memory on Dsub due to its high space cost.}
\label{main-results}
\begin{tabular}{c|c|cccc}
\toprule
\multirow{2}{*}{\textbf{Backbone}} & \multirow{2}{*}{\textbf{Method}} & \textbf{Wikipedia} & \textbf{Reddit} & \textbf{Dsub} & \textbf{CoOAG} \\ \cline{3-6}
 & &AUC & AUC & AUC & ACC  \\
\midrule
\multirow{6}{*}{TGAT} 
 & CFT & $77.43 (\pm 3.01)$ & $82.68 (\pm 0.06)$ & $62.32 (\pm 1.27)$ & $86.28 (\pm 0.18)$ \\
 & DLS & $79.56 (\pm 2.55)$ & $78.66 (\pm 0.04)$ & -- & -  \\
 & NPL & $78.52 (\pm 2.28)$ & $80.72 (\pm 2.65)$ & $61.71 (\pm 2.21)$ & $87.67 (\pm 0.61)$ \\
 & \ourmethod{}-2D & $78.20 (\pm 6.83)$ & $85.84 (\pm 6.09)$ & $60.76 (\pm 2.91)$ & $88.37 (\pm 0.16)$ \\
 & SEM & $\underline{81.09} (\pm 3.62)$ & $\underline{86.27} (\pm 5.99)$ & $\underline{64.34} (\pm 0.99)$ & $\underline{88.38} (\pm 0.38)$ \\
 & \textbf{Ours} & $\mathbf{85.52} (\pm 3.29)$ & $\mathbf{87.31} (\pm 6.50)$ & $\mathbf{65.07} (\pm 1.57)$ & $\mathbf{89.05} (\pm 0.63)$ \\ \midrule
 
\multirow{6}{*}{TCL} 
 & CFT & $76.27 (\pm 4.68)$ & $84.48 (\pm 5.53)$ & $62.60 (\pm 1.26)$ & $86.12 (\pm 1.11)$ \\
 & DLS & $80.55 (\pm 1.93)$ & $82.85 (\pm 0.38)$ & -- & -- \\
 & NPL & $77.71 (\pm 5.66)$ & $84.20 (\pm 3.86)$ & $63.59 (\pm 3.09)$ & $87.59 (\pm 0.26)$ \\
 & \ourmethod{}-2D & $76.68 (\pm 2.86)$ & $86.98 (\pm 3.34)$ & $60.64 (\pm 1.41)$ & $\underline{87.94} (\pm 0.30)$ \\
 & SEM & $\underline{81.02} (\pm 2.82)$ & $\underline{87.56} (\pm 1.56)$ & $\underline{65.11} (\pm 1.26)$ & $87.90 (\pm 0.18)$ \\
 & \textbf{Ours} & $\mathbf{82.27} (\pm 4.62)$ & $\mathbf{89.41} (\pm 3.32)$ & $\mathbf{66.80} (\pm 2.45)$ & $\mathbf{88.24} (\pm 0.26)$ \\ \midrule

\multirow{6}{*}{TGN}
 & CFT & $80.68 (\pm 2.02)$ &  $\mathbf{89.69} (\pm 2.07) $ & OOM & $83.65 (\pm 0.63)$ \\
 & DLS & $78.48 (\pm 1.60)$ & $ 80.92 (\pm 4.99)$ & -- & -- \\
 & NPL & $\underline{87.58} (\pm 2.14)$ & $84.22 (\pm 2.48)$ & OOM & $86.13(\pm0.31)$ \\
 & \ourmethod{}-2D & $86.59 (\pm 3.01)$ & $\underline{86.76} (\pm 3.89)$ & OOM & $\underline{86.23}(\pm0.66)$ \\
 & SEM & $86.34 (\pm 2.66)$ & $82.61 (\pm 3.07)$ & OOM & $86.07 (\pm0.66)$ \\
 & \textbf{Ours} & $\mathbf{87.97} (\pm 2.90)$ & $84.32 (\pm 2.07)$ & OOM & $\mathbf{86.71} (\pm 0.66)$ \\ \midrule

\multirow{6}{*}{GraphMixer} 
 & CFT & $76.60 (\pm 2.00)$ & $66.11 (\pm 6.04)$ & $62.78 (\pm 1.90)$& $85.63 (\pm 0.14)$ \\
 & DLS & $80.70 (\pm 4.00)$ & $61.97 (\pm 7.36)$ & -- &  -- \\
 & NPL & $80.86 (\pm 1.62)$ & $\underline{71.72} (\pm 6.48)$ & $67.14 (\pm 1.68)$ & $86.98 (\pm 0.61)$ \\
 & \ourmethod{}-2D & $81.41 (\pm 4.25)$ & $66.86 (\pm 11.14)$ & $62.33 (\pm 1.35)$ & $87.51 (\pm 0.51)$ \\
 & SEM & $\underline{83.33} (\pm 1.45)$ & $68.65 (\pm 3.70)$ & $\underline{69.23} (\pm 1.92)$ & $\underline{88.07} (\pm 0.30)$ \\
 & \textbf{Ours} & $\mathbf{84.09} (\pm 0.95)$ & $\mathbf{71.93} (\pm 7.94)$ & $\mathbf{69.76} (\pm 1.54)$ & $\mathbf{88.26} (\pm 0.38)$ \\ \midrule

\multirow{6}{*}{DyGFormer}
 & CFT & $64.76 (\pm 9.21)$ & $67.14 (\pm 8.04)$ & $68.48 (\pm 1.46)$ & $85.27 (\pm 0.83)$ \\
 & DLS & $71.95 (\pm 2.29)$ & $64.63 (\pm 4.90)$ & -- & -- \\
 & NPL & $\underline{73.85} (\pm 5.44)$ & $67.44 (\pm 3.47)$ & $\underline{70.31} (\pm 1.11)$ & $\underline{86.16} (\pm0.38)$ \\
 & \ourmethod{}-2D & $66.48 (\pm 6.76)$ & $71.14 (\pm6.27)$ & $69.11 (\pm 2.96)$ & $86.04(\pm0.30)$ \\
 & SEM & $70.91 (\pm 8.80)$ & $\underline{71.59}(\pm4.51)$ & $69.75(\pm2.47)$ & $86.07 (\pm 0.66)$ \\
 & \textbf{Ours} & $\mathbf{74.85} (\pm 3.07)$ & $\mathbf{75.86} (\pm 8.04)$ & $\mathbf{72.39} (\pm 1.91)$ & $\mathbf{86.26} (\pm 0.27)$ \\ 
 \bottomrule
\end{tabular}
\end{table*}

\section{Experiments}

Our experiments are designed to address the following key research questions (RQs):

\textbf{RQ1}: How does \ourmethod{} perform compared to other baselines when evaluated on the final timestamp labels?

\textbf{RQ2}: Does pseudo-labels generated by \ourmethod{} improve performance by capturing the dynamic information of nodes? 

\textbf{RQ3}: Does the proposed Temporal Curriculum Learning strategy improve performance? 

\textbf{RQ4}: Is \ourmethod{} stable and computationally efficient?

\subsection{Experiment Settings}

\subsubsection{Datasets} 
We evaluate \ourmethod{} in various real-world scenes. Specifically, we use three existing datasets: \textit{Wikipedia} \cite{kumar2019predicting}, \textit{Reddit} \cite{kumar2019predicting} and a sub-graph of Dgraph \cite{huang2022dgraph} (\textit{Dsub} for short). Moreover, based on the OAG dataset \cite{sinhaOverviewMicrosoftAcademic2015,zhangOAGLinkingLargescale2019,zhangOAGLinkingEntities2023,tangArnetMinerExtractionMining2008}, we propose a new dataset \textit{CoOAG}. Statistics and more details of datasets can be found in Appendix.\ref{Dataset Details}. For Wikipedia, Reddit, and Dsub (imbalanced binary classification tasks), we use AUC as the evaluation metric \cite{yu2023towards}, excluding background nodes in Dsub. For CoOAG (multi-class classification tasks), we use accuracy (ACC) as the metric.
To align with real-world scenarios where only the final labels are available for evaluation, we design a new dataset split and evaluation protocol. Specifically, we split the datasets based on the final timestamp and evaluate model performance only on the final timestamp labels ($\mathcal{Y}_{F,A}$). By statistically analyzing the labels at the final timestamp, we partition the nodes into training, validation, and test sets with a ratio of 7:1.5:1.5 \cite{yu2023towards}. All experiments are conducted with five random seeds, and metrics are reported as the average performance. Implementation details and hyperparameter settings can be found in Appendix.\ref{Implementation Details} and Appendix.\ref{Hyperparameters}. 

\begin{figure}
    \centering
    \includegraphics[width=\linewidth]{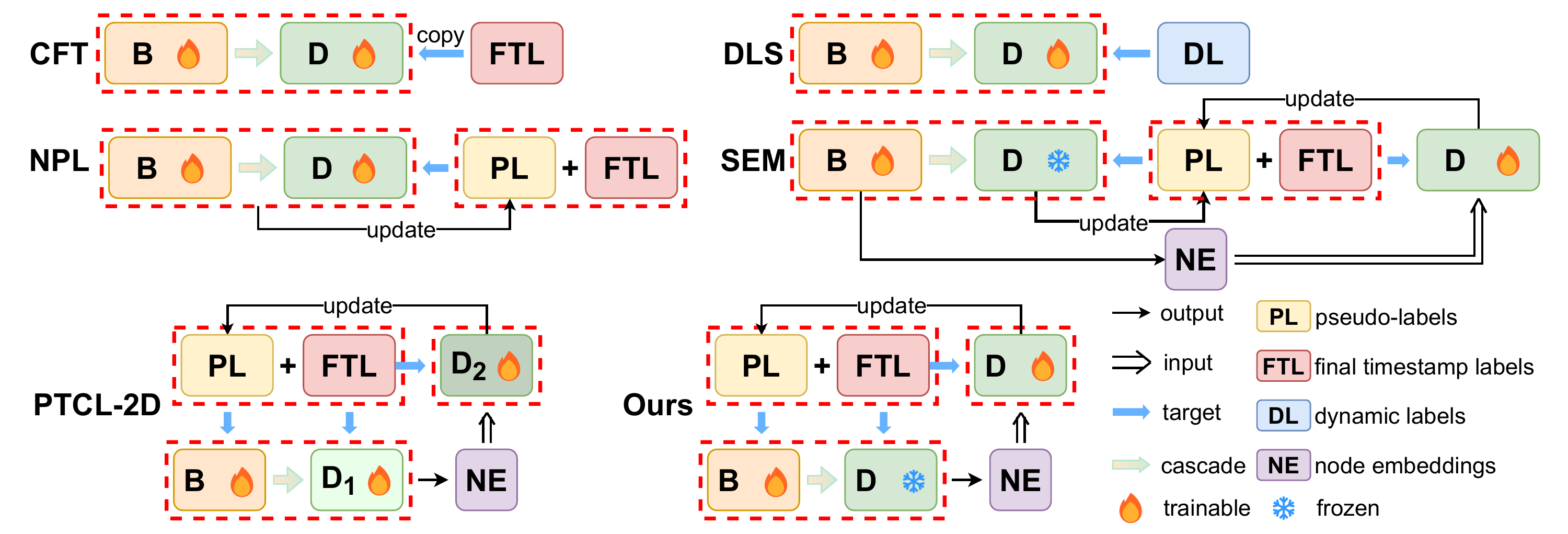}
    \caption{Architecture of different baselines and \ourmethod{}. "B" stands for backbone, and "D" stands for decoder.}
    \label{fig:baseline}
\end{figure}

\vspace{-5pt}

\subsubsection{Baselines} 
We compare \ourmethod{} with several different methods that can be adapted to our task. And specifically, we use five different dynamic backbones as backbones: TGAT \cite{xu2020inductive}, GraphMixer \cite{cong2023we}, TCL \cite{Wang2021TCLTD}, TGN \cite{rossi2020temporal}, and DyGFormer \cite{yu2023towards}, and apply a simple MLP as the decoder. More backbone details are in Appendix.\ref{backbone details}. As shown in Figure \ref{fig:baseline}, the baselines are designed to cover a range of approaches:

\begin{itemize}[left=0pt..1em, topsep=0pt, itemsep=-1pt, parsep=0pt]
\item \textbf{CFT (Copy-Final Timestamp labels)}: A naive baseline that simply copies the final timestamp labels ($\mathcal{Y}_{F,B}$) to earlier timestamps as approximations of dynamic labels ($\mathcal{Y}_{E,B}$) for training.
\item \textbf{DLS (Dynamic Label Supervision)}: A baseline that performs supervised training directly using the dynamic labels provided by the dataset (Wikipedia, Reddit), where available \cite{yu2023towards, rossi2020temporal}.
\item \textbf{NPL (Naive Pseudo-Labels)}: A variant of \ourmethod{} that uses pseudo-labels but jointly optimizes the backbone and decoder, without EM optimization.
\item \textbf{\ourmethod{}-2D (\ourmethod{} with 2 Decoders)}: A variant of \ourmethod{} that uses two decoders: one decoder is trained exclusively on the final timestamp labels (E-step), generating pseudo-labels, while the other decoder is jointly optimized with the backbone on weighted pseudo-labels and final timestamp labels (M-step). The final embeddings are provided by the backbone for the E-step training.
\item \textbf{SEM (Standard EM)}: A variant of \ourmethod{} where both the E-step and M-step are trained on the weighted loss of pseudo-labels and final timestamp labels \cite{Zhao2022LearningOL}, while other components remain the same as \ourmethod{}. In this way, E-step uses Eq. (\ref{e-objective1}) as the objective function instead of Eq. (\ref{e-objective2}).
\end{itemize}

\subsection{RQ1: Main Results}
To evaluate the effectiveness of \ourmethod{}, we conduct experiments by replacing different backbones and comparing various baselines. The results in Table \ref{main-results} show that \ourmethod{} improves the performance of nearly all the backbones with respect to all datasets.

\subsubsection{Superiority of Pseudo-Labels} 
\ourmethod{} consistently outperforms the CFT and DLS baselines with significant improvements  (ranging from a minimum of 0.99$\%$ to a maximum of 11.23$\%$ in AUC/ACC), demonstrating the effectiveness of using pseudo-labels. Compared to simply copying final timestamp labels (CFT), our approach dynamically learns pseudo-labels, effectively mitigating the issue of incorrect information propagation caused by direct label copying. Notably, \ourmethod{} even surpasses supervised training with dynamic labels (DLS), suggesting that the pseudo-labels generated by \ourmethod{} better reflect the underlying node information changes than the original dynamic labels. We provide a detailed analysis of this phenomenon in Section \ref{ps-label analysis}.

\subsubsection{Necessity of Separate Optimization} \label{Necessity of Separate Optimization}
Compared to the NPL baseline, which jointly optimizes the backbone and decoder, \ourmethod{} achieves significantly better performance with an average improvement of 2.74$\%$ in AUC/ACC. This highlights the importance of separate optimization, which ensures that the decoder remains consistent with the final timestamp labels while allowing the backbone to effectively learn from pseudo-labels at earlier timestamps. Additionally, \ourmethod{} outperforms SEM, demonstrating that training the decoder exclusively on final timestamp labels ($\alpha =0$, as described in Section \ref{e-step}) leads to better results.

\subsubsection{Advantages of \ourmethod{} over \ourmethod{}-2D} 
\ourmethod{} not only achieves better performance but also requires fewer computational resources compared to the \ourmethod{}-2D baseline. This indicates that using a single decoder helps maintain consistency in the feature space between the backbone and decoder, leading to more stable training and higher performance.

\subsection{RQ2: Pseudo-Label Analysis} \label{ps-label analysis}

In this section, we evaluate the effectiveness of our pseudo-labels and their ability to capture dynamic patterns through two experiments.

\begin{table}[t]
\centering
\caption{AUC comparison of different backbones using Dynamic Label Supervised Learning (DLS) and Pseudo-Label Supervised Learning (PLS) on Wikipedia Dataset.}
\label{tab:pseudo_label_analysis}
\begin{tabular}{c|ccccc}
\toprule
  & TGAT & TCL & TGN & GraphMixer & DyGFormer\\
\midrule
DLS     &  79.56 & 80.55 & 78.48 & 80.7 & 71.95\\
\midrule
PLS     &  81.11 & 82.19 & 79.32 & 81.02 & 72.42 \\
\toprule
\end{tabular}
\end{table}

\begin{figure}[t]
\centering
\includegraphics[width=1.0\linewidth]{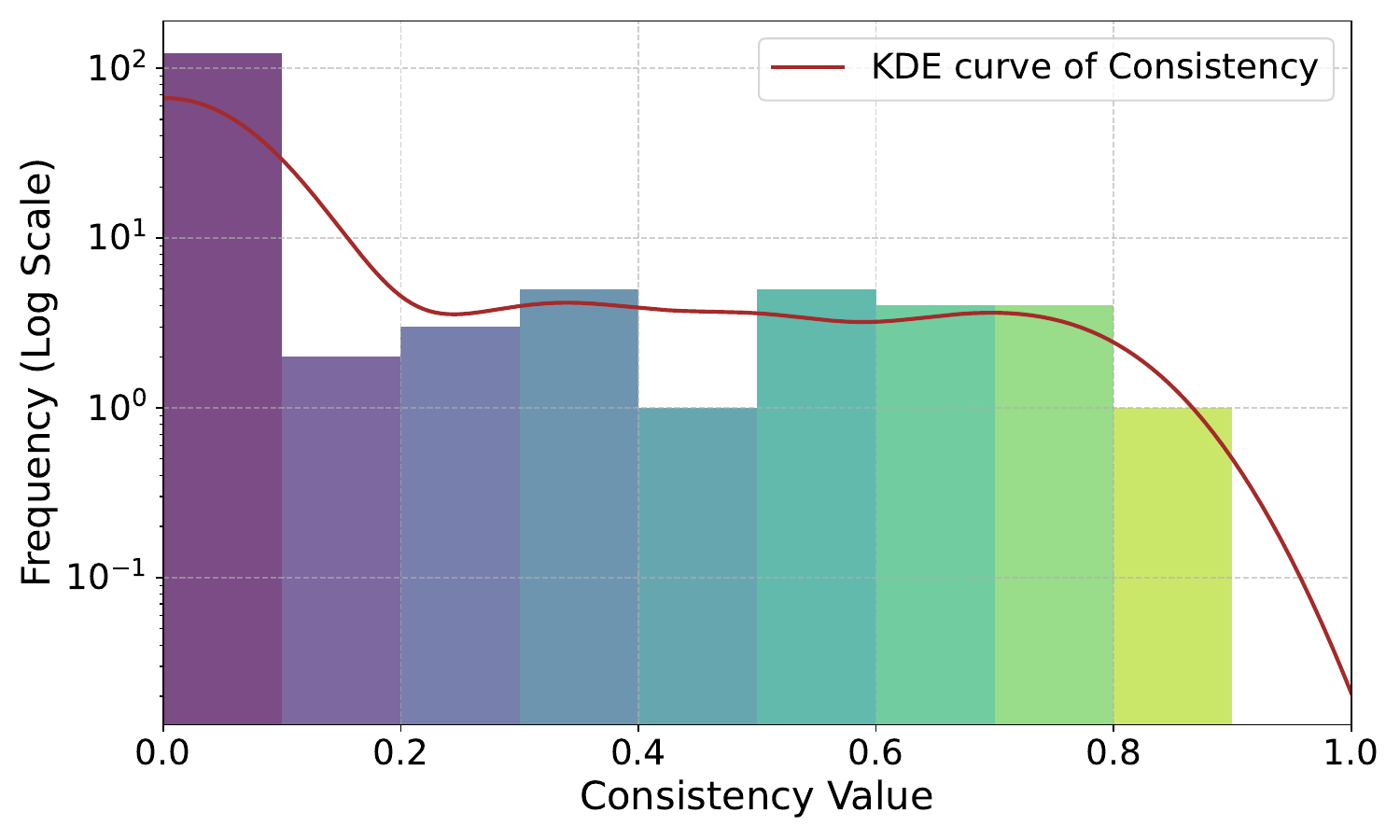}
\caption{Histogram of pseudo-labels consistency.}
\label{fig:ps}
\vspace{-10pt}
\end{figure}

\subsubsection{Pseudo-label Supervision Study}
To analyze the effectiveness of our pseudo-labels, we conduct the following experiment: We train models from scratch using our generated pseudo-labels (from our trained model) as full supervision labels and compare the results with DLS. As shown in Table \ref{tab:pseudo_label_analysis}, models trained with our pseudo-labels consistently outperform those using original dynamic labels with an average improvement of 0.96$\%$ in AUC, demonstrating our pseudo-labels better capture latent evolutionary patterns of nodes.

\subsubsection{Label Consistency Analysis} 
To further investigate the temporal changes of labels, we analyze the labels' consistency on Wikipedia's positive samples. Consistency is quantified as follows:
\begin{equation}
\begin{aligned}
    \hat{N}_{u'} = \max \Big\{ k \in \mathbb{N}^+ \mid & y_{u'}^{t_i} = y_{u'}^{T_{u'}}, \\
    & \forall i \in \{|\mathcal{T}_{u'}|-k,\dots,|\mathcal{T}_{u'}|-1\} \Big\} 
\end{aligned}
\end{equation}
\begin{equation}
    C_{u'} = \frac{\hat{N}_{u'}}{|\mathcal{T}_{u'}| - 1}
\end{equation}
where $u' \in \mathcal{V}_{\text{neg}}$, $\mathcal{V}_{\text{neg}} = \{u' | u' \in \mathcal{V},y_{u'}^{T_{u'}} = 1\}$. In Wikipedia, the dynamic label consistency of all negative samples is zero($C_{u'} \equiv 0$), which means that negative-class labels abruptly emerge at the final timestamp ($\mathcal{Y}_F$), neglecting gradual behavioral transitions. And CFT method enforces strict label continuity by replicating $\mathcal{Y}_{F,B}$($C_{u'} \equiv 1$), suffering from severe feature misalignment. In contrast, as shown in Figure \ref{fig:ps}, pseudo-labels generated by \ourmethod{} with TGAT backbone exhibit varying consistency values, propagating smoothly across timestamps in $\mathcal{Y}_{E,B}$, striking an optimal balance between temporal coherence and feature alignment.

% This advantage stems from two key design choices: (1) \textbf{Temporal Decoupling}: \ourmethod{} separates the backbone from the decoder, allowing the backbone to capture evolutionary patterns without overfitting to sparse final labels. (2) \textbf{Curriculum Learning Weighting}: By prioritizing pseudo-labels near the final timestamp, \ourmethod{} ensures high confidence while filtering out noisy predictions from earlier stages.

\subsection{RQ3: Temporal Curriculum Learning Analysis}
To comprehensively evaluate our Temporal Curriculum Learning design, we conduct a comparison experiment against the naive solution, which uses all the generated pseudo-labels, and two commonly used baseline strategies to choose more reliable pseudo-labels in Curriculum Learning, as introduced in Section \ref{Related Work Pseudo-Labeling}:
\begin{itemize}[left=0pt..1em, topsep=0pt, itemsep=-1pt, parsep=0pt]
\item \textit{Confidence Score Threshold (CST)} \cite{Sun2019MultiStageSL,He2022HowDP,CascanteBonilla2020CurriculumLR}: This method filters pseudo-labels based on their confidence scores, improving the overall quality of the labels.
\item \textit{Entropy of Softmax Trajectory (EST)} \cite{Song2019SELFIERU,Pei2024MemoryDA}: This method filters pseudo-labels using the entropy of the softmax trajectory, which is an accumulated distribution that summarizes the model's disagreement across training rounds.
\end{itemize}

\begin{table}[t]
    \centering
    \caption{AUC comparison of our Temporal Curriculum Learning with two commonly used strategies to choose more reliable pseudo-labels in Curriculum Learning and a naive strategy which uses all the pseudo-labels. CST stands for Confidence Score Threshold, and EST stands for Entropy of Softmax Trajectory. Bold indicates the best performance, \underline{underline} the second best.}
    \begin{tabular}{c|c|cccc}
        \toprule
        Backbone & Dataset & Naive & CST & EST & Ours \\
        \midrule
        \multirow{2}{*}{TGAT} 
          & Wikipedia & \underline{82.38} & 79.48 & 81.89 & \textbf{85.52} \\ %done
          & Dsub      & \underline{64.12} & 62.78 & 63.80 & \textbf{65.07} \\ %done
        \midrule
        \multirow{2}{*}{TCL}  
          & Wikipedia & \underline{81.77} & 78.06 & 80.56 & \textbf{82.27} \\ %done
          & Dsub       & \underline{65.29} & 63.68 & 64.57 & \textbf{66.80} \\ %done
        \midrule
        \multirow{2}{*}{TGN}  
          & Wikipedia & 86.33 & 83.63 & \underline{86.88} & \textbf{87.97} \\ %done
          & Dsub       & OOM & OOM & OOM & OOM \\ %done
        \midrule
        \multirow{2}{*}{GraphMixer} 
          & Wikipedia & \underline{83.34} & 81.59 & 81.13 & \textbf{84.09} \\ %done
          & Dsub       & 68.08 & \underline{68.79} & 67.55 & \textbf{69.76} \\ %done
        \midrule
        \multirow{2}{*}{DyGFormer} 
          & Wikipedia & \underline{69.42} & 67.15 & 68.39 & \textbf{74.85} \\ %done
          & Dsub       & \underline{72.30} & 70.86 & 69.21 & \textbf{72.39} \\ %done
        \bottomrule
    \end{tabular}
    \label{tab:ablation}
\end{table}

As shown in Table \ref{tab:ablation}, it's obvious that our Temporal Curriculum Learning strategy consistently achieves the highest AUC scores across all backbones and datasets, indicating the effectiveness of our design. The Naive strategy performs significantly worse than \ourmethod{}, with an average AUC drop of 1.74$\%$, highlighting the noise introduced by relying on the pseudo-labels. Similarly, CST and EST, which rely on static filtering mechanisms, mostly underperform both \ourmethod{} and the Naive strategy due to their inability to adapt to the dynamic reliability of pseudo-labels. This comparison demonstrates that our Temporal Curriculum Learning design, by effectively utilizing temporal information, is better suited for the task.

% The superiority of our approach stems from its ability to dynamically assign higher weights to pseudo-labels closer to the final timestamps, effectively filtering noise from earlier steps while still leveraging early-stage information by dynamically adjusting the weight distribution across different timestamps. In contrast, both CST and EST employ static filtering mechanisms that fail to adapt to the evolving reliability of pseudo-labels throughout the temporal sequence. Our Temporal Curriculum Learning ensures optimal utilization of temporal information while maintaining robustness against noisy early predictions.

\begin{figure}[t]
    \centering
    \includegraphics[width=1.0\linewidth]{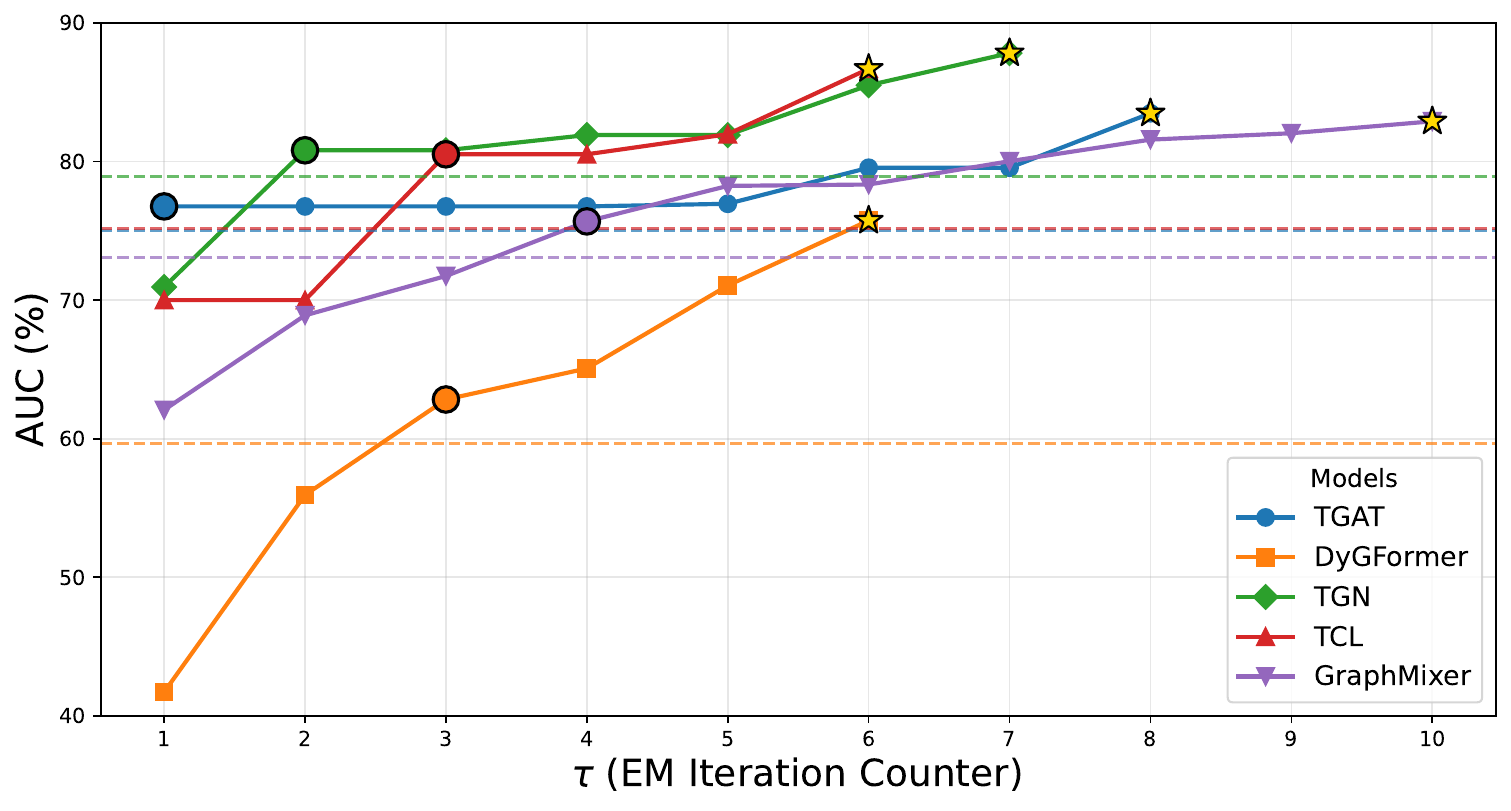}
    \caption{Convergence curves for 5 backbones. Star markers ($\star$) denote peak performance; circled points($\bullet$)indicate surpassing baselines. Dashed lines show baseline AUC.}
    \label{fig:convergence}
    % \vspace{-16pt}
\end{figure}
% \vspace{-15pt}

\subsection{RQ4: Convergence and Efficiency Analysis}
We validate the convergence behavior and computational efficiency of \ourmethod{} through iterative training analysis on the Wikipedia dataset with 5 backbones. The compared baseline is the CFT, which only utilizes the $\mathcal{Y}_F$. 
As shown in Figure \ref{fig:convergence}, \ourmethod{} achieves rapid convergence across all backbones on Wikipedia. TGAT surpasses its baseline immediately at the first iteration, while others require 2–4 iterations. All models reach peak performance within 6–10 iterations, with DyGFormer achieving the largest gain (+16.1 $\%$ AUC). Despite iterative training, each EM iteration incurs 0.8$\times$ – 1.2$\times$ time overhead compared to baseline training, making the total cost practical.  This confirms \ourmethod{} achieves robust convergence across diverse backbones while maintaining computational practicality.

\section{Conclusion}

In this work, we bridge the gap between highly dynamically changing graphs and limited accessible labels. Specifically, we propose \ourmethod{}, an extensible method for dynamic node classification. By generating temporally-weighted pseudo-labels and optimizing through the variational EM framework, our method achieves significant improvements (up to 11.23$\%$ AUC/ACC gain) over baselines in various real-world scenes, including financial and academic networks, etc., which subsequently validates the importance of tracking dynamic historical changes as well as the effectiveness of the variational EM framework and Temporal Curriculum Learning. The introduced CoOAG dataset and FLiD framework enable a practical evaluation of temporal label propagation. Notably, \ourmethod{} also can be applied to other dynamic graph tasks due to its high dynamic nature capturing capabilities. To sum up, this work establishes foundational techniques for learning evolving node behaviors under realistic annotation constraints, laying a solid foundation for future dynamic graph studies.


% \begin{acks}
% \end{acks}

\bibliographystyle{ACM-Reference-Format}
\bibliography{reference}

\clearpage
\appendix

\section{Algorithm for Training \ourmethod{}}
\label{algorithm}

\begin{algorithm}[h]
\caption{Optimization Algorithm}
\label{alg:optimization}
\begin{algorithmic}[1]
    \State \textbf{Input:} A dynamic graph $\mathcal{G}$, final timestamp labels $\mathcal{Y}_{F,B}$, and hyperparameter $\beta$
    \State \textbf{Output:} Predicted $\hat{\mathcal{Y}}_{F,A}$.
    
    \State $\theta \gets \arg\min_{\theta} \mathcal{L}_{lp}$ \Comment{Warm up the backbone}
    \State $\tau \gets 1$  \Comment{Initialize iteration counter}
    
    \Repeat        
        \State \textbf{E-Step: Decoder Optimization}
        \State \hspace{1em} $\phi \gets \arg\max_{\phi} \mathcal{O}_\phi$ \Comment{Update the decoder $q_{\phi}$ }
        \State \hspace{1em} $\hat{\mathcal{Y}}_{E,B} \gets \arg\max q_{\phi}(\mathcal{Y}_{E,B} | \mathcal{G})$ \Comment{Generate pseudo-labels}

        \State \textbf{M-Step: Backbone Optimization}
        \State \hspace{1em} $w_u^{t,\tau} \gets f_{\text{TW}}(t, u, \tau, T_u, \gamma)$ \Comment{Compute temporal weights}
        \State \hspace{1em} $\theta \gets \arg\max_{\theta} \mathcal{O}_\theta$ \Comment{Update backbone $p_{\theta}$}

        \State $\tau \gets \tau + 1$ \Comment{Update iteration counter}
    
    \Until Converged

    \State $\hat{\mathcal{Y}}_{F,A} \gets \arg\max q_{\phi}(\mathcal{Y}_{F,A} | \mathcal{G})$ \Comment{Final prediction}

    \State \textbf{return} $\hat{\mathcal{Y}}_{F,A}$
\end{algorithmic}
\end{algorithm}

\section{Dataset Details}
\label{Dataset Details}
Due to the lack of widely studied datasets for the \textbf{label-limited dynamic node classification}, we utilize three existing datasets that align closely with our task and propose a new dataset \textit{CoOAG} specifically designed for this problem. The statistics of four datasets are introduced in Table \ref{tab:datasets_all}.

\begin{table}[h]
    \centering    
    \caption{Dataset Statistics}
    \label{tab:datasets_all}
    \begin{tabular}{c|cccc}
        \toprule
         &  Wikipedia & Reddit & Dsub & CoOAG \\
         \midrule
        Nodes & $9,227$ & $10,984$ & $150,000$ & $9,559$ \\
        Edges & $15,7474$ & $67,2447$ & $16,8154$ & $11,4337$\\
        Duration & 1 month & 1 month & 1 year & 22 years \\
        Total classes & $2$ & $2$ & $2$ & $5$\\
        Bipartite & \Checkmark & \Checkmark & \XSolidBrush & \XSolidBrush \\
        Node Feat Dim & -- & -- & $34$ & $384$\\
        Edge Feat Dim & 172 & 172 & $1$ & $384$ \\
        \bottomrule
        
    \end{tabular}

\end{table}

\subsection{Previous datasets}

We use three publicly available datasets and do the preprocessing to adapt them to our task:

\subsubsection{Dsub}

   \textbf{Description}: Dsub is a subgraph of the Dgraph \cite{huang2022dgraph} dataset, which is a financial fraud detection dataset where nodes represent users, and edges represent emergency contact relationships between users. Node labels indicate whether a user is ultimately identified as fraudulent (failing to repay loans over an extended period). Node features are derived from user metadata. In addition to confirmed fraudulent and non-fraudulent labels, the dataset includes background nodes that lack sufficient information for labeling but are retained to maintain graph connectivity.

   \textbf{Preprocessing}: To facilitate efficient training, we extract a subgraph called Dsub using Breadth-First Search (BFS), ensuring that the subgraph remains connected and preserves the original label distribution.
   
\subsubsection{Wikipedia}

   \textbf{Description}: Wikipedia \cite{kumar2019predicting} is a bipartite interaction graph that records edits on Wikipedia pages over one month. Nodes represent users and pages, and edges denote editing behaviors with timestamps. Each edge is associated with a 172-dimensional Linguistic Inquiry and Word Count (LIWC) feature. The dataset includes dynamic labels indicating whether users are temporarily banned from editing \cite{yu2023towards}.

   \textbf{Task Adaptation}: To simulate real-world scenarios where only the final labels are available, we split the dynamic labels into $\mathcal{Y}_{F,B}$ (final labels) and $\mathcal{Y}_{E,B}$ (unobserved labels). During training, only $\mathcal{Y}_{F,B}$ is used.
   
\subsubsection{Reddit}

   \textbf{Description}: Reddit \cite{kumar2019predicting} is a bipartite graph that records user posts under subreddits over one month. Nodes represent users and subreddits, and edges represent timestamped posting requests. Each edge is associated with a 172-dimensional LIWC feature. The dataset includes dynamic labels indicating whether users are banned from posting \cite{yu2023towards}.
   
   \textbf{Task Adaptation}: Similar to Wikipedia, we split the dynamic labels into $\mathcal{Y}_{F,B}$ and $\mathcal{Y}_{E,B}$, using only $\mathcal{Y}_{F,B}$ for training to simulate real-world conditions.

\subsection{CoOAG}

\begin{table}[t]
\centering
\caption{Label Distributions of CoOAG}
\label{tab:dataset_stats}
\begin{tabular}{lcc}
\toprule
\textbf{Label Distribution} & \\
\midrule
\quad  ROB & 2,845 (29.64\%) \\
\quad  CV & 1,700 (17.71\%) \\
\quad  NLP & 1,652 (17.21\%) \\
\quad  AI/ML & 1,971 (20.53\%) \\
\quad  DM/WS & 1,431 (14.91\%) \\
\bottomrule
\end{tabular}
\end{table}

% \subsubsection{CoarXiv}

%    \textbf{Description}: arXiv is an academic collaboration network where nodes represent authors and edges represent co-authorship relationships based on papers published on arXiv. Node labels are assigned based on authors' research interests, focusing on the four primary categories: CV, NLP, ROB, and AI/ML.
   
%     \textbf{Preprocessing}: We utilize the arXiv Dataset \cite{clement2019arxiv} with multi-stage filtering:
%     \begin{enumerate}
%         \item Retain authors with $\geq$ 20 publications in target categories;
%         \item Label authors by majority voting of their last 5 papers;
%         \item Construct edge features by concatenating paper metadata and abstracts, encoded via \texttt{all-MiniLM-L12-v2}\cite{SentencetransformersAllMiniLML12v2Hugging2024}; 
%         \item Generate node features through mean pooling of associated paper embeddings. Edge timestamps are assigned using paper last-update dates.
%     \end{enumerate}

\subsubsection{Description}: To advance research in this domain, we introduce CoOAG, a novel dataset derived from the academic sphere, inspired by the Coauthor CS and Coauthor Physics networks \cite{shchurPitfallsGraphNeural2019}. This dataset has undergone stringent quality control and temporal consistency checks. Label distributions are detailed in Table~\ref{tab:dataset_stats}.

The CoOAG dataset is constructed using the Microsoft Academic Graph (MAG) portion from Open Academic Graph 2.1\cite{sinhaOverviewMicrosoftAcademic2015,zhangOAGLinkingLargescale2019,zhangOAGLinkingEntities2023,tangArnetMinerExtractionMining2008}, with a focus on publications from leading AI conferences. The node labels in CoOAG denote authors' research interests, classified into the following categories:

\begin{itemize}[left=0pt..1em, topsep=0pt, itemsep=-1pt, parsep=0pt]
    \item CV (Computer Vision)
    \item NLP (Natural Language Processing)
    \item ROB (Robotics)
    \item DM/WS (Data Mining/Web Search)
    \item AI/ML (Other AI Fields)
\end{itemize}

\subsubsection{Preprocessing}: We employ structured prompts with the Qwen-Plus API \cite{qwen2,qwen2.5} to categorize research domains using paper Fields of Study (FoS) and abstracts. The prompt template, as illustrated in Listing~\ref{lst:classification_prompt}, encompasses:
\begin{itemize}[left=0pt..1em, topsep=0pt, itemsep=-1pt, parsep=0pt]
    \item Category definitions with canonical examples
    \item Strict output format constraints
    \item Weighted keyword matching logic
    \item Interactive classification examples
\end{itemize}
This approach achieves 98.3\% ACC on 120 manually verified samples. Edge features are generated by concatenating paper metadata and abstracts, encoded using the all-MiniLM-L12-v2 model. Node features are computed as the average of all paper features for each author. Conference submission deadlines determine edge timestamps. The classification workflow maintains temporal consistency by processing papers in chronological order.

\begin{tcolorbox}[label={box:classification_prompt}, colback=blue!5!white,colframe=blue!65!white,title=Research Domain Classification Prompt Template]
\begin{lstlisting}[label={lst:classification_prompt}, basicstyle=\small\ttfamily, frame=tb]
Classify the author's research domain into one of the following 5 categories based on the given field keywords and weights:
- 0: CV (Computer Vision)
- 1: NLP (Natural Language Processing)
- 2: ROB (Robotics)
- 3: DM/WS (Data Mining/Web Search)
- 4: AI/ML (Other AI Fields)

Input: Multiple field keywords with weights
Output requirements:
    - **(*@\textbf{Format}@*)**: Directly return classification result (0-4)
    - **(*@\textbf{Constraint}@*)**: Answer must be a single digit without explanation

Examples:
    - Input: "[computer vision (0.53377)] [image filter (0.5337)]"
    - Output: 0
    
Input: 
    {fos_text}
\end{lstlisting}
\end{tcolorbox}

\section{Backbone Details}
\label{backbone details}
\begin{itemize}[left=0pt..1em, topsep=0pt, itemsep=-1pt, parsep=0pt]
    \item \textbf{TGAT} \cite{xu2020inductive} leverages a self-attention mechanism to simultaneously capture spatial and temporal dependencies. Initially, TGAT combines the raw node feature $\mathbf{n}_u$ with a learnable time encoding $z(t)$, forming $\mathbf{n}_u(t) = [\mathbf{n}_u || z(t)]$, where $z(t) = \cos(tw + b)$. Subsequently, self-attention is applied to generate the representation of node $u$ at time $t_0$, denoted as $\mathbf{h}_u^{t_0} = \text{SAM}(\mathbf{n}_u(t_0), \{\mathbf{n}_v(m_v) \mid v \in N_{t_0}(u)\})$. Here, $N_{t_0}(u)$ represents the set of neighbors of node $u$ at time $t_0$, and $m_v$ indicates the timestamp of the most recent interaction involving node $v$. Finally, predictions for any node pair $(u, v)$ at time $t_0$ are obtained via $\text{MLP}([\mathbf{h}_u^{t_0} || \mathbf{h}_v^{t_0}])$.

    \item \textbf{TCL} \cite{Wang2021TCLTD} adopts a contrastive learning framework. To construct interaction sequences for each node, TCL employs a breadth-first search algorithm on the temporal dependency subgraph. A graph transformer is then utilized to learn node representations by jointly considering graph topology and temporal dynamics. Additionally, TCL integrates a cross-attention mechanism to model the interdependencies between interacting nodes.

    \item \textbf{TGN} \cite{rossi2020temporal} combines RNN-based and self-attention-based techniques. TGN maintains a memory module to store and update the state $s_u(t)$ of each node $u$, which serves as a compact representation of $u$'s historical interactions. Given the memory updater as mem, when an edge $e_{uv}(t)$ connecting nodes $u$ and $v$ is observed, the memory state of node $u$ is updated as $s_u(t) = \text{mem}(s_u(t^-), s_v(t^-) || \mathbf{e}_{u,v}^t)$, where $s_u(t^-)$ denotes the memory state of $u$ just prior to time $t$, and $\mathbf{e}_{u,v}^t$ represents the edge feature. The memory updater $\text{mem}$ is implemented using a recurrent neural network (RNN). Node embeddings $\mathbf{h}_u^t$ are computed by aggregating information from the $L$-hop temporal neighborhood through self-attention.

    \item \textbf{GraphMixer} \cite{cong2023we} introduces a simple yet effective MLP-based architecture. Instead of relying on trainable time encodings, GraphMixer utilizes a fixed time encoding function, which is integrated into a link encoder based on MLP-Mixer to process temporal links. A node encoder with neighbor mean-pooling is employed to aggregate node features. Specifically, for each node $u$, GraphMixer computes its embedding $\mathbf{h}_u^t$ by summarizing the features of its neighbors within the temporal context.

    \item \textbf{DyGFormer} \cite{yu2023towards} employs a self-attention mechanism to model dynamic graphs. For a given node $u$, DyGFormer retrieves the features of its involved neighbors and edges to represent their encodings. It incorporates a neighbor co-occurrence encoding scheme, which captures the frequency of each neighbor's appearance in the interaction sequences of both the source and destination nodes, thereby explicitly exploring pairwise correlations. Rather than operating at the interaction level, DyGFormer divides the interaction sequences of each source or destination node into multiple patches, which are then processed by a transformer to compute node embeddings $\mathbf{h}_u^t$.
\end{itemize}

\section{Implementation Details}
\label{Implementation Details}
We use PyTorch \cite{paszke2019pytorch}, scikit-learn \cite{pedregosa2011scikit}, PyTorch Geometric \cite{Fey/Lenssen/2019}, DyGLib \cite{yu2023towards} library to implement our proposed framework FLiD.
We conduct experiments on two clusters: (1) 4×Tesla V100 (32GB memory) using 16-core CPUs and 395GB RAM; (2) 8×2080Ti (11GB memory) using 12-core CPUs and 396GB RAM.

\section{Hyperparameters}
We optimize all methods across all models using the Adam optimizer \cite{Kingma2014AdamAM}, with cross-entropy loss as the objective function. Initially, we warm up all backbones through link prediction tasks \cite{kumar2019predicting}. Subsequently, the entire models are trained for 100 epochs, employing an early stopping strategy with a patience of 15. For consistency, we set the learning rate to 0.0001 and the batch size to 200 across all methods and datasets. To ensure robustness and minimize deviations, we conduct five independent runs for each method with random seeds ranging from 0 to 4 and report the average performance \cite{yu2023towards}. 

\label{Hyperparameters}
\subsection{Model Configurations}
Here, we present the configurations for each model (Table \ref{tab:configs}): TGAT, TGN, TCL, GraphMixer, and DyGFormer, all of which remain consistent across datasets.

\begin{table}[htbp]
    \centering
    \caption{Model Configurations Comparison}
    \label{tab:configs}
    \resizebox{0.47\textwidth}{!}{ % 调整表格宽度
    \begin{tabular}{l c c c c c}
        \toprule
        \textbf{Hyperparameter} & \textbf{TGAT} & \textbf{TGN} & \textbf{TCL} & \textbf{GraphMixer} & \textbf{DyGFormer} \\
        \midrule
        Time encoding dim & 100 & 100 & 100 & 100 & 100 \\
        Output dim & 172 & 172 & 172 & 172 & 172 \\
        Attention heads & 2 & 2 & 2 & -- & 2 \\
        \addlinespace
        Graph conv layers & 2 & 1 & -- & -- & -- \\
        Transformer layers & -- & -- & 2 & -- & 2 \\
        MLP-Mixer layers & -- & -- & -- & 2 & -- \\
        \addlinespace
        Node memory dim & -- & 172 & -- & -- & -- \\
        Depth encoding dim & -- & -- & 172 & -- & -- \\
        Co-occurrence dim & -- & -- & -- & -- & 50 \\
        Aligned encoding dim & -- & -- & -- & -- & 50 \\
        \addlinespace
        Memory updater & -- & GRU & -- & -- & -- \\
        Time gap $T$ & -- & -- & -- & 2000 & -- \\
        \bottomrule
    \end{tabular}
    }
\end{table}

\subsection{\ourmethod{} Hyperparameters}
Here, we present the hyperparameters of \ourmethod{} (Table \ref{tab:PTCL_hyper}): $\beta$ is a hyperparameter that balances the weight of pseudo-labels and final timestamp labels; $\gamma$ is a hyper-parameter that controls the rate of Temporal Curriculum fdecay. Note that TGN runs out of memory on Dsub due to its high space cost.

\begin{table}[htbp]
    \centering
    \caption{Hyperparameters of \ourmethod{}}
    \label{tab:PTCL_hyper}
    \resizebox{0.47\textwidth}{!}{ % 调整表格宽度
        \begin{tabular}{c|c|cccc}
        \toprule
        \textbf{Model} & \textbf{Hyperparameters} & \textbf{Wikipedia} & \textbf{Reddit} & \textbf{Dsub} & \textbf{CoOAG} \\
        \midrule
        \multirow{2}{*}{TGAT} 
        
         & $\beta $ & $0.9$ & $0.9$ & $0.7$ & $0.2$ \\
         & $\gamma $ & $0.8$ & $0.1$ & $0.2$ & $0.9$  \\
         \midrule
         
        \multirow{2}{*}{TCL} 
        
         & $\beta $ & $0.1$ & $0.9$ & $0.1$ & $0.8$ \\
         & $\gamma $ & $0.6$ & $0.9$ & $0.5$ & $0.6$  \\
         \midrule
         
        \multirow{2}{*}{TGN} 
        
         & $\beta $ & $0.9$ & $0.9$ & -- & $0.9$ \\
         & $\gamma $ & $0.05$ & $0.01$ & -- & $0.1$  \\
         \midrule
         \multirow{2}{*}{GraphMixer} 
        
         & $\beta $ & $0.5$ & $0.5$ & $0.5$ & $0.5$ \\
         & $\gamma $ & $1.3$ & $0.1$ & $0.1$ & $0.4$  \\
         \midrule
         
         \multirow{2}{*}{DyGFormer} 
        
         & $\beta $ & $0.7$ & $0.5$ & $0.1$ & $0.1$ \\
         & $\gamma $ & $0.01$ & $0.01$ & $0.5$ & $0.1$  \\
         \bottomrule
        \end{tabular}
    }
\end{table}

\end{document}